\documentclass{article}

\usepackage{microtype}
\usepackage{graphicx}
\usepackage{subcaption}
\usepackage{booktabs} 
\usepackage{multirow}
\usepackage{hyperref}

\usepackage[preprint]{icml2026}

\usepackage{amsmath}
\usepackage{amssymb}
\usepackage{mathtools}
\usepackage{amsthm}
\usepackage{bm}
\usepackage[table]{xcolor}

\usepackage[capitalize,noabbrev]{cleveref}

\theoremstyle{plain}

\theoremstyle{definition}

\theoremstyle{remark}

\usepackage[textsize=tiny]{todonotes}

\icmltitlerunning{A Unified Spatial Alignment Framework for Highly Transferable Transformation-Based Attacks on Spatially Structured Tasks}

\begin{document}

\definecolor{mdarkred}{RGB}{0, 110, 0}
\definecolor{red}{RGB}{255, 0, 0}

\twocolumn[
  \icmltitle{A Unified Spatial Alignment Framework for Highly Transferable Transformation-Based Attacks on Spatially Structured Tasks}



  \icmlsetsymbol{equal}{*}

  \begin{icmlauthorlist}
    \icmlauthor{Jiaming Liang}{sch}
    \icmlauthor{Chi-Man Pun}{sch}
  \end{icmlauthorlist}

  \icmlaffiliation{sch}{University of Macau}

  \icmlcorrespondingauthor{Chi-Man Pun}{cmpun@um.edu.mo}

  \icmlkeywords{Machine Learning, ICML}

  \vskip 0.3in
]



\printAffiliationsAndNotice{}  

\begin{abstract}
Transformation-based adversarial attacks (TAAs) demonstrate strong transferability when deceiving classification models. However, existing TAAs often perform unsatisfactorily or even fail when applied to structured tasks such as semantic segmentation and object detection. Encouragingly, recent studies that categorize transformations into non-spatial and spatial transformations inspire us to address this challenge. We find that for non-structured tasks, labels are spatially non-structured, and thus TAAs are not required to adjust labels when applying spatial transformations. In contrast, for structured tasks, labels are spatially structured, and failing to transform labels synchronously with inputs can cause spatial misalignment and yield erroneous gradients. To address these issues, we propose a novel unified \textit{Spatial Alignment Framework (SAF)} for highly transferable TAAs on spatially structured tasks,
where the TAAs spatially transform labels synchronously with the input using the proposed \textit{Spatial Alignment (SA)} algorithm.  Extensive experiments demonstrate the crucial role of our SAF for TAAs on structured tasks. Specifically, in non-targeted attacks, our SAF degrades the average mIoU on Cityscapes from $24.50$ to $11.34$, and on Kvasir-SEG from $49.91$ to $31.80$, while reducing the average mAP of COCO from $17.89$ to $5.25$. 
\end{abstract}

\section{Introduction}
\begin{figure}[t]
    \centering
    \includegraphics[width=0.99\linewidth]{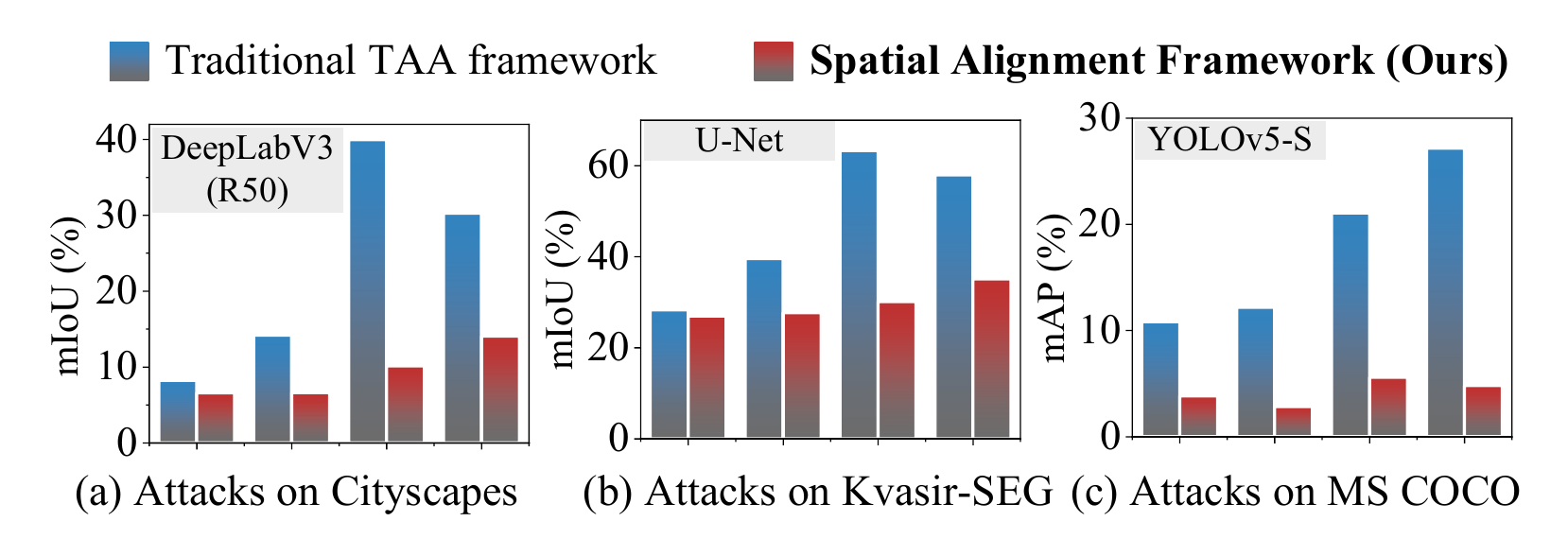}
    \caption{Performance gains from integrating spatial alignment framework into various TAAs across spatially structured tasks. Four groups of bars correspond to DEM, SIA, BSR, and I-C, respectively. Surrogate models are indicated in gray boxes.}
    \label{fig:teaser}
\end{figure}

Adversarial attacks~\cite{goodfellow2014explaining} exploit imperceptible perturbations to provoke substantial deviations in model behavior. To boost attack efficacy, diverse strategies have been proposed. Among them, transformation-based attacks (TAAs)~\cite{xie2019improving} create multiple perturbation counterparts from randomly parameterized input transformations and average them to avoid local optima, thereby improving transferability. Benefiting from being end-to-end, training-free, high-speed, and highly transferable, TAAs have become a promising branch.

Unfortunately, current TAAs~\cite{lin2024boosting, zhu2024learning, wang2024boosting, liang2025ic, guo2025boosting} are limited to attacking non-structured tasks\footnote{In this paper, "structured tasks" is a shorthand for tasks with spatially structured prediction~\cite{tsochantaridis2005large}.} (e.g., classification), perform significantly worse on structured tasks, such as semantic segmentation and object detection, and may nearly fail in some cases (e.g., targeted attacks). This has been a long-standing open problem.

TASS~\cite{he2024transferable} directly applies TAAs to semantic segmentation but perform unsatisfactorily. OSFD~\cite{ding2024transferable}, a bold attempt combining TAAs with intermediate feature-based attacks, achieved remarkable results in attacking object detection models. However, this approach compromises the end-to-end efficiency of TAAs and exacerbates implementation complexity. Moreover, stricter constraints, such as smaller transformation magnitudes, pose obstacles to directly applying existing TAAs to structured tasks. Importantly, the reasons why current end-to-end TAAs fail in structured tasks remain unanswered. Therefore, this study aims to investigate the differences in TAAs between structured and non-structured tasks and seeks to establish a unified framework to extend existing end-to-end TAAs to structured tasks preserving their original design. 

The recent decomposition of transformations in TAAs into non-spatial and spatial transformations~\cite{liang2025ic} motivates our solution to this challenge. 
Specifically, to improve transferability, most advanced TAAs integrate spatial transformations. In non-structured tasks, labels are spatially non-structured and thus need not be synchronously transformed. In contrast, structured tasks involve spatially structured labels, and applying spatial transformations without synchronizing the labels introduces spatial misalignment, leading to spatially inconsistent gradients. To address this, we propose \textit{Spatial Alignment (SA)}, which synchronously transforms structured labels with the input when applying TAAs to structured tasks, thereby eliminating spatial misalignment and enabling TAAs. We refer to the unified TAA framework integrated with SA for structured tasks as the \textit{Spatial Alignment Framework (SAF)}.


We evaluate SAF in combination with various TAAs across diverse structured tasks. Remarkably, for non-targeted attacks, SA reduces the average mIoU from $24.50$ to $11.34$ on Cityscapes and from $49.91$ to $31.80$ on Kvasir-SEG, while degrading the average mAP of MS COCO from $17.89$ to $5.25$. Preview of the results is presented in Figure~\ref{fig:teaser}. The main contributions are summarized as follows:

\begin{itemize}

    \item We reveal that TAAs fail in structured tasks due to spatial misalignment. To address this, we propose a novel \textit{Spatial Alignment (SA)} algorithm, which synchronously transforms labels with inputs, and integrate SA into existing TAA pipelines to form a unified \textit{Spatial Alignment Framework (SAF)} for structured tasks.

    \item Extensive experiments on semantic segmentation and object detection demonstrate that our SAF substantially improves the performance of TAAs on structured tasks. On Cityscapes, SAF-based TAAs exhibit superior transferability, surpassing state-of-the-art attacks.

    \item To our knowledge, this is the first work to successfully apply end-to-end TAAs to structured tasks. Extensive results show that TAAs behave differently on structured tasks than on non-structured tasks, suggesting the need to reconsider TAAs in structured scenarios.
\end{itemize}

\section{Related Work}
\subsection{Adversarial Attacks for Structured Tasks}

\textbf{Adversarial Attacks for Semantic Segmentation.} DAG~\cite{xie2017adversarial} is the first work to extend adversarial examples to semantic segmentation, with an initial study of their transferability. To improve the efficiency of adversarial example generation and enhance attack effectiveness, SegPGD~\cite{gu2022segpgd} adapts PGD to semantic segmentation, establishing a strong baseline. TranSegPGD~\cite{jia2023transegpgd} proposes a two-stage strategy to balance the attack on hard-to-attack pixels and the transferability of adversarial examples. CosPGD~\cite{agnihotri2024cospgd} proposes an alignment score that promotes globally balanced perturbations instead of localized pixel-wise distortions, thereby improving effectiveness. RP-PGD~\cite{zhang2025rp} integrates region-based and prototype-based strategies to improve performance.

\textbf{Adversarial Attacks for Object Detection.}
DAG~\cite{xie2017adversarial} is also the first work to extend adversarial examples to object detection. Subsequently, RAP~\cite{li2018robust} and CAP~\cite{zhang2020contextual} were proposed to attack two-stage object detectors. To improve transferability, CATA~\cite{cai2022context} generates adversarial examples by leveraging the co-occurrence of objects along with their relative positions and sizes as contextual information. To bypass context-consistency checks, ZQTA~\cite{cai2022zero} adopts a zero-query setting and generates context-consistent attack plans, selecting the most effective one via a perturbation success probability matrix. OSFD~\cite{ding2024transferable} integrates transformation-based and intermediate-feature-based attacks, enhancing adversarial transferability.

Despite significant progress in transfer attacks for classification tasks, their counterparts in structured prediction tasks have received little attention. Techniques that are widely effective for classification, including transformation-based, intermediate-feature-based, ensemble-based, and generative strategies, remain largely unexplored in structured tasks. In this paper, we address this gap by exploring the application of transformation-based strategies in structured prediction.

\begin{figure*}[t]
    \centering
    \includegraphics[width=0.98\linewidth]{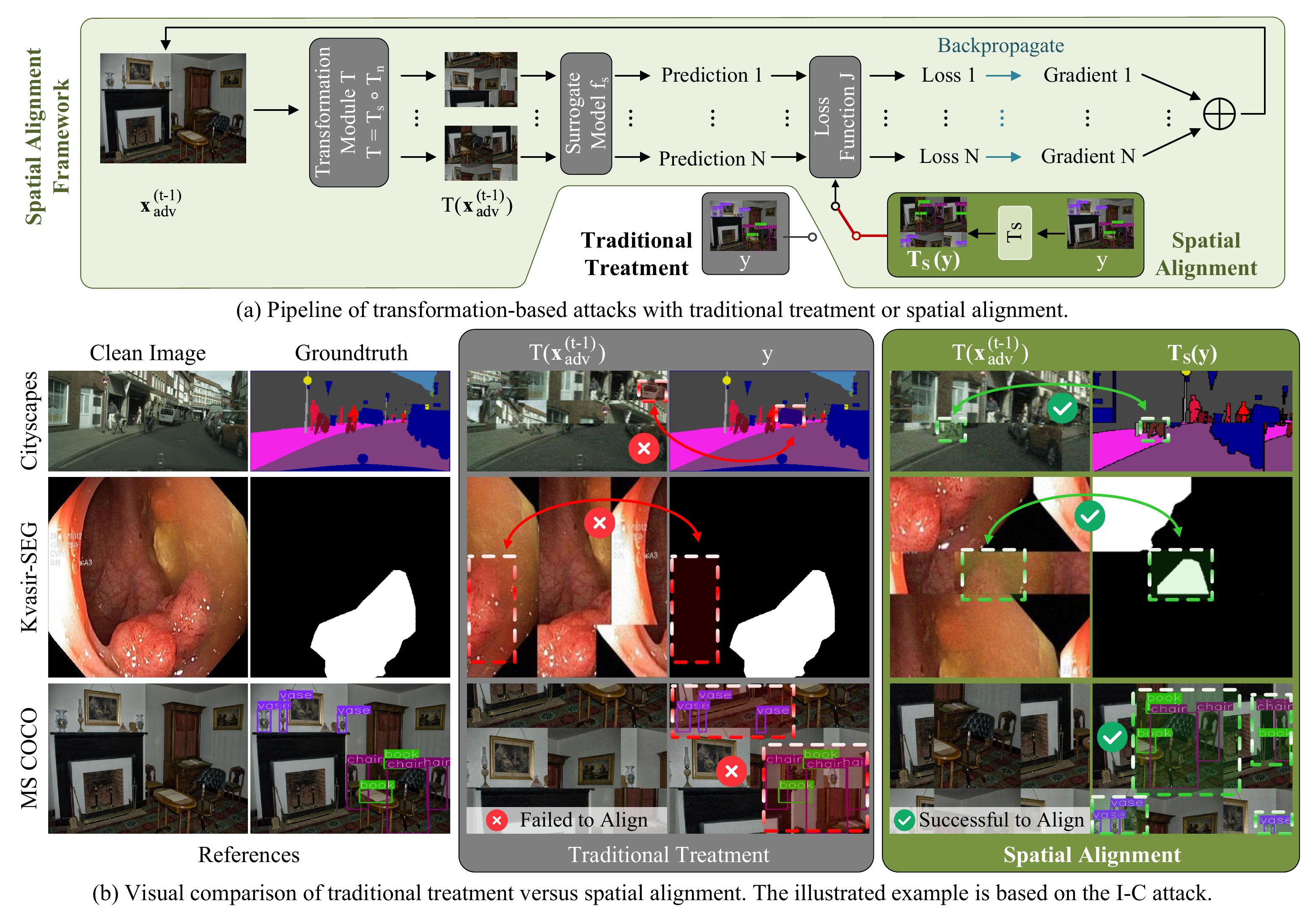}
    \caption{Illustration of the proposed \textit{Spatial Alignment Framework (SAF)}.}
    \label{fig:illustration_misalignment_alignment}
\end{figure*}

\subsection{Transformation-based Adversarial Attacks}
\textbf{Overview.} Transformation-based attacks leverage transformation modules to augment the surrogate model, preventing adversarial examples from being overfitted to the surrogate and thereby improving transferability. Prior work has explored various transformations to enhance transferability in classification tasks. DIM~\cite{xie2019improving}, TIM~\cite{dong2019evading}, and SIM~\cite{lin2019nesterov} employ resizing, translation, and scaling as transformation modules, respectively. Admix~\cite{wang2021admix} employs mixup as its transformation. ATTA~\cite{wu2021improving} applies transformations via an adversarial transformation network. SSIM~\cite{long2022frequency} inject noise in the frequency domain as its transformation. STM~\cite{ge2023improving} adopts a style transfer network as the transformation module. DeCoWA~\cite{lin2024boosting} uses image warping as its transformation. BSR~\cite{wang2024boosting} cascades block shuffling and rotation as transformations. I-C~\cite{liang2025ic} identifies checkerboard artifacts in adversarial examples and proposes a transformation involving cascading noise addition, block shuffling, and resizing. AITL~\cite{yuan2022adaptive}, SIA~\cite{wang2023structure}, L2T~\cite{zhu2024learning}, and OPS~\cite{guo2025boosting} integrate multiple transformations to further enhance transferability.

\textbf{Bold Attempts at Transformation-Based Attacks on Structured Tasks.}
While the application of TAAs in classification tasks has been widely explored, their use in structured tasks remains underexplored, with only a few bold attempts at preliminary research. OSFD~\cite{ding2024transferable} applies transformation-based strategies to object detection attacks, but only in combination with intermediate-feature-based methods. In contrast, conventional TAAs operate end-to-end without accessing intermediate features. Moreover, OSFD strictly limits transformation magnitude (e.g., rotations up to $7^\circ$ and resizing up to $1.1$), whereas traditional TAAs rely on larger transformations to achieve high transferability. Consequently, many existing TAAs cannot be directly applied to object detection models. TASS~\cite{he2024transferable} directly applies TAAs to semantic segmentation models. However, it only conducts preliminary explorations based on DIM and TIM. Moreover, their results indicate that naively applying TAAs to semantic segmentation models in a conventional manner may even degrade attack performance, sometimes performing worse than not using transformations. Therefore, we aim to investigate the key reasons why TAAs cannot be directly applied to spatially structured tasks and to propose a solution for the future exploration of the potential of TAAs in structured tasks.

\section{Methodology}

\subsection{Preliminaries}
\label{section:Preliminaries}
\textbf{Problem Formulation.} Given image–label pairs $(\bm{x}, \bm{y})\in (\bm{\mathcal{X}}, \bm{\mathcal{Y}})$, under the $p$-ball constraint $B_p(\bm{x},\epsilon)=\{\bm{x}'\in \bm{\mathcal{X}} \mid ||\bm{x}'-\bm{x}||_p\leq \epsilon\}$ with the perturbation budget $\epsilon$, we aim to find an attack $A\in\mathcal{A}$ such that the generated adversarial example $\bm{x}_{adv}=A(\bm{x},f_{s})$ by the surrogate $f_{s}$ satisfies
\begin{equation}
\begin{aligned}
    \min \Sigma_{(\bm{x},\bm{y})\in(\bm{\mathcal{X}}, \bm{\mathcal{Y}})} \text{criterion}(f_{t}(\bm{x}_{adv}), \bm{y}),
\end{aligned}
\label{equation:definition_attacks}
\end{equation}
where $f_{t}$ indicates the target model, and the function $\text{criterion}(\cdot)$ indicates the objective of the attack. Following prior work, this paper focuses on the $l_{\infty}$ constraint. For classification tasks, the criterion is a $\{0,1\}$-valued function that takes the value $1$ if and only if 
\begin{equation}
\begin{aligned}
    \arg\max f_{t}(\bm{x}_{adv})=\arg\max\bm{y}.
\end{aligned}
\label{equation:definition_classification_criterion}
\end{equation}
For semantic segmentation tasks, the criterion is typically the IoU or its variants, such as mIoU. For object detection tasks, the criterion is usually the mAP.

\textbf{Framework of TAAs.} TAAs have proven to be a promising attack subspace for classification tasks. Based on MI-FGSM, TAAs can be described within the following framework:
\begin{equation}
\begin{aligned}
    \bar{\bm{\mu}}=\frac{1}{N}\Sigma_{i=1}^{N}\nabla_{\bm{x}_{adv}^{(t-1)}}J(f_{s}(T(\bm{x}_{adv}^{(t-1)})),\bm{y}),
\end{aligned}
\label{equation:definition_TAA_1}
\end{equation}
\begin{equation}
\begin{aligned}
    \bm{g}^{(t)}=\lambda\cdot \bm{g}^{(t-1)}+\frac{\bar{\bm{\mu}}}{||\bar{\bm{\mu}}||_{1}},
\end{aligned}
\label{equation:definition_TAA_2}
\end{equation}
\begin{equation}
\begin{aligned}
    \bm{x}_{adv}^{(t)}=\bm{x}_{adv}^{(t-1)}+\alpha\cdot \operatorname{sgn}(\bm{g}^{(t)}).
\end{aligned}
\label{equation:definition_TAA_3}
\end{equation}
Here, $T$ denotes a transformation with random parameters, $J$ is the loss function, $N$ is the number of transformation counterparts, $\lambda$ is the momentum coefficient, $\alpha$ is the perturbation step size, and $\operatorname{sgn}(\cdot)$ denotes the sign function.

\subsection{Causes of TAAs Failure in Structured Tasks}
\label{section:Causes}
Existing end-to-end TAAs are typically tailored for classification tasks. To improve transferability, these TAAs often combine multiple types of transformations, including non-spatial transformations and spatial transformations~\cite{liang2025ic}. Among them, spatial transformations alter the spatial structure of images. Empirical evidence shows that they are key contributors to transferability and are therefore widely adopted by existing TAAs. Nevertheless, the label $\bm{y}$ in classification tasks is a spatially non-structured vector. Since the use of spatial transformations in these TAAs does not alter the underlying image semantics, the label $\bm{y}$ does not require synchronized modification with the spatial transformations. As shown in Equation~\ref{equation:definition_TAA_1}, current TAAs adopt this strategy, where the $t$-round input $\bm{x}_{adv}^{(t-1)}$ is transformed by $T$ while the label $\bm{y}$ remains unchanged.

Unfortunately, unlike the spatially non-structured labels in classification tasks, the labels in spatially structured tasks, such as semantic segmentation maps and object detection bounding boxes, are inherently spatially structured. Handling spatial transformations in structured tasks in the same manner as described in Equation~\ref{equation:definition_TAA_1} leads to catastrophic errors, which constitutes a fundamental reason why many TAAs fail on spatially structured tasks.

To further illustrate the impact of spatial transformations in TAAs on structured tasks, we decompose the transformation $T$ in Equation~\ref{equation:definition_TAA_1} into non-spatial transformations $T_{n}$ and spatial transformations $T_{s}$, with $\circ$ denoting the composition of transformations. Equation~\ref{equation:definition_TAA_1} can then be rewritten as 
\begin{equation}
\begin{aligned}
    \bar{\bm{\mu}}=\frac{1}{N}\Sigma_{i=1}^{N}\nabla_{\bm{x}_{adv}^{(t-1)}}J(f_{s}(T_{s}\circ T_{n}(\bm{x}_{adv}^{(t-1)})),\bm{y}).
\end{aligned}
\label{equation:definition_TAA_decomposition}
\end{equation}
For structured tasks, the prediction on a spatially transformed input $f(T_{s}(\bm{x}))$ can be approximated by applying the same spatial transformation to the prediction of the input $T_{s}(f(\bm{x}))$. Therefore, Equation~\ref{equation:definition_TAA_decomposition} can be approximated as
\begin{equation}
\begin{aligned}
    \bar{\bm{\mu}}\simeq\frac{1}{N}\Sigma_{i=1}^{N}\nabla_{\bm{x}_{adv}^{(t-1)}}J(T_{s}(f_{s}(T_{n}(\bm{x}_{adv}^{(t-1)}))),\bm{y}).
\end{aligned}
\label{equation:definition_TAA_decomposition_rewritten}
\end{equation}

\begin{algorithm}[t]
\caption{Spatial Alignment for Structured Tasks}  
\label{algorithm: Spatial Alignment}  
\begin{algorithmic}[1]

\REQUIRE{Image-label pair $(\bm{x},\bm{y})$, surrogate model $f_{s}$, loss function $J$, attack $A$ with transformations $T_{s}$ and $T_{n}$.
\REQUIRE Perturbation budget $\epsilon$, the number of iterations $L$, perturbation step size $\alpha$, momentum coefficient $\lambda$, the number of transformation counterparts $N$.}
\ENSURE{L-round adversarial example $\bm{x}_{adv}^{(L)}$.}

\STATE $\bm{x}_{adv}^{(0)}=\bm{x}$, $\bm{g}^{(0)}=\bm{0}$

\FOR{$t=1$ to $L$}

\STATE $\bar{\bm{\mu}}^{(t)}=\bm{0}$
\FOR{$k=1$ to $N$}

\STATE $\nabla_{k}^{(t)}=\nabla_{\bm{x}_{adv}^{(t-1)}}J(f_{s}(T(\bm{x}_{adv}^{(t-1)})),\mathcolor{mdarkred}{T_{s}(\bm{y})})$

\STATE $\bar{\bm{\mu}}^{(t)}=\bar{\bm{\mu}}^{(t)}+\nabla_{k}^{(t)}/N$

\ENDFOR

\STATE $\bm{g}^{(t)}=\lambda\cdot \bm{g}^{(t-1)}+\bar{\bm{\mu}}^{(t)}/||\bar{\bm{\mu}}^{(t)}||_{1}$

\STATE $\bm{x}_{adv}^{(t)}=\bm{x}_{adv}^{(t-1)}+\alpha\cdot \operatorname{sgn}(\bm{g}^{(t)})$

\ENDFOR

\STATE \textbf{Return} $\bm{x}_{adv}^{(L)}$
\end{algorithmic}

\end{algorithm}

In Equation~\ref{equation:definition_TAA_decomposition_rewritten}, the label $\bm{y}$ actually corresponds to the prediction $f_{s}(T_{n}(\bm{x}_{adv}^{(t-1)}))$. Since $f_{s}(T_{n}(\bm{x}_{adv}^{(t-1)}))$ is further transformed by spatial transformations $T_{s}$, the label $\bm{y}$ requires the synchronized spatial transformation for alignment to obtain correct gradient information. However, existing TAAs do not apply synchronized transformations to $\bm{y}$.

The visualization in Figure~\ref{fig:illustration_misalignment_alignment}(b) intuitively illustrates this issue. Taking a sample from Kvasir-SEG and I-C attack as an illustration, the block shuffle served as $T_{s}$ in I-C attack divides the image into rearranged blocks, causing the distribution of the polyps in $T(\bm{x}_{adv}^{(t-1)})$ to no longer align with the label $\bm{y}$. However, I-C attack still computes gradients with the label $\bm{y}$, leading to errors due to this misalignment.

\newcommand{\tabincell}[2]{\begin{tabular}{@{}#1@{}}#2\end{tabular}}
\begin{table*}[t]
    \centering
    
    \setlength{\tabcolsep}{1.4mm}
    \caption{mIoU (\%) of non-targeted attacks on Cityscapes. * indicates white-box attacks. -SA denotes attacks with spatial alignment. The better result between attacks with and without spatial alignment is highlighted in \textbf{bold}.}
    \fontsize{9}{10}\selectfont
    \begin{tabular}{cllcccccccccc}
         \toprule
         \multirow{2}{*}[0ex]{Surrogate}&\multirow{2}{*}[0ex]{Attack}&\multirow{2}{*}[0ex]{Venue}&\multicolumn{1}{c}{DLV3}&\multicolumn{1}{c}{DLV3}&\multicolumn{1}{c}{DLV3+}&\multicolumn{1}{c}{FPN}&\multicolumn{1}{c}{PSPNet}&\multicolumn{1}{c}{UN++}&\multicolumn{1}{c}{UPerN}&\multicolumn{1}{c}{DPT}&\multicolumn{1}{c}{SegFormer}&\multirow{2}{*}[0ex]{AVG} \\
         &&&\multicolumn{1}{c}{R50}&\multicolumn{1}{c}{EB0}&\multicolumn{1}{c}{R101}&\multicolumn{1}{c}{R50}&\multicolumn{1}{c}{D161}&\multicolumn{1}{c}{Xcept}&\multicolumn{1}{c}{VGG16}&\multicolumn{1}{c}{V-B/16}&\multicolumn{1}{c}{MiT-B2}& \\
         \midrule
         
         \multicolumn{3}{c}{Clean}&68.42&62.79&69.36&66.08&59.50&68.20&66.51&62.98&71.94&66.20 \\
         \cmidrule{1-13}

        \multirow{11}{*}[-2ex]{\tabincell{c}{DLv3\\R50}}
        &SegPGD&ECCV'22&0.84*&43.43&34.25&6.91&17.41&43.57&33.90&53.79&51.46&31.73 \\
        &CosPGD&ICML'24&0.06*&45.58&37.67&9.40&17.85&45.92&34.41&54.87&54.85&33.40 \\
        \cmidrule{2-13}
        &DEM&ECCV'20&1.46*&11.80&6.32&3.90&2.03&7.19&3.33&22.19&16.02&8.25 \\
        &\cellcolor{gray!20}DEM-SA&\cellcolor{gray!20}Ours&\cellcolor{gray!20}\textbf{0.91}*&\cellcolor{gray!20}\textbf{9.90}&\cellcolor{gray!20}\textbf{4.34}&\cellcolor{gray!20}\textbf{3.04}&\cellcolor{gray!20}\textbf{1.52}&\cellcolor{gray!20}\textbf{5.77}&\cellcolor{gray!20}\textbf{2.67}&\cellcolor{gray!20}\textbf{19.29}&\cellcolor{gray!20}\textbf{11.89}&\cellcolor{gray!20}\textbf{6.59} \\
        \cmidrule{2-13}
         &SIA&ICCV'23&6.52*&17.49&12.45&8.41&6.16&14.05&9.89&32.05&20.80&14.20 \\
         &\cellcolor{gray!20}SIA-SA&\cellcolor{gray!20}Ours&\cellcolor{gray!20}\textbf{1.81}*&\cellcolor{gray!20}\textbf{8.24}&\cellcolor{gray!20}\textbf{4.26}&\cellcolor{gray!20}\textbf{2.64}&\cellcolor{gray!20}\textbf{1.99}&\cellcolor{gray!20}\textbf{5.18}&\cellcolor{gray!20}\textbf{3.68}&\cellcolor{gray!20}\textbf{23.33}&\cellcolor{gray!20}\textbf{8.34}&\cellcolor{gray!20}\textbf{6.61} \\
         \cmidrule{2-13}

        &BSR&CVPR'24&38.62*&41.05&43.40&40.84&31.04&41.91&31.54&45.86&45.35&39.96 \\
        &\cellcolor{gray!20}BSR-SA&\cellcolor{gray!20}Ours&\cellcolor{gray!20}\textbf{3.90}*&\cellcolor{gray!20}\textbf{13.85}&\cellcolor{gray!20}\textbf{9.66}&\cellcolor{gray!20}\textbf{5.29}&\cellcolor{gray!20}\textbf{4.33}&\cellcolor{gray!20}\textbf{7.85}&\cellcolor{gray!20}\textbf{5.48}&\cellcolor{gray!20}\textbf{26.06}&\cellcolor{gray!20}\textbf{14.81}&\cellcolor{gray!20}\textbf{10.14} \\
        \cmidrule{2-13}

        &I-C&MM'25&26.05*&33.41&32.66&28.08&18.68&30.79&24.12&41.59&37.22&30.29 \\
        &\cellcolor{gray!20}I-C-SA&\cellcolor{gray!20}Ours&\cellcolor{gray!20}\textbf{6.27}*&\cellcolor{gray!20}\textbf{18.09}&\cellcolor{gray!20}\textbf{14.71}&\cellcolor{gray!20}\textbf{8.29}&\cellcolor{gray!20}\textbf{5.28}&\cellcolor{gray!20}\textbf{12.33}&\cellcolor{gray!20}\textbf{11.51}&\cellcolor{gray!20}\textbf{30.68}&\cellcolor{gray!20}\textbf{19.41}&\cellcolor{gray!20}\textbf{14.06} \\
        \midrule

        \multirow{8}{*}[-2ex]{\tabincell{c}{SegFormer\\MiT-B2}}
        &SegPGD&ECCV'22&42.01&41.93&45.32&38.81&20.47&38.47&23.80&48.22&0.63*&33.30 \\
        &CosPGD&ICML'24&44.57&44.53&47.98&41.89&22.56&42.70&27.64&50.08&0.05*&35.78 \\
        \cmidrule{2-13}
        &DEM&ECCV'20&11.95&18.18&17.91&16.17&5.50&11.63&5.00&26.47&2.41*&12.80 \\
        &\cellcolor{gray!20}DEM-SA&\cellcolor{gray!20}Ours&\cellcolor{gray!20}\textbf{8.25}&\cellcolor{gray!20}\textbf{14.69}&\cellcolor{gray!20}\textbf{13.33}&\cellcolor{gray!20}\textbf{11.74}&\cellcolor{gray!20}\textbf{3.90}&\cellcolor{gray!20}\textbf{8.91}&\cellcolor{gray!20}\textbf{3.74}&\cellcolor{gray!20}\textbf{21.17}&\cellcolor{gray!20}\textbf{1.54}*&\cellcolor{gray!20}\textbf{9.70} \\
        \cmidrule{2-13}
        
         &SIA&ICCV'23&21.56&24.23&26.10&23.99&10.41&20.64&12.59&34.11&7.82*&20.16 \\
         &\cellcolor{gray!20}SIA-SA&\cellcolor{gray!20}Ours&\cellcolor{gray!20}\textbf{9.27}&\cellcolor{gray!20}\textbf{12.95}&\cellcolor{gray!20}\textbf{13.32}&\cellcolor{gray!20}\textbf{10.95}&\cellcolor{gray!20}\textbf{3.42}&\cellcolor{gray!20}\textbf{9.32}&\cellcolor{gray!20}\textbf{4.84}&\cellcolor{gray!20}\textbf{24.14}&\cellcolor{gray!20}\textbf{1.29}*&\cellcolor{gray!20}\textbf{9.94} \\
         \cmidrule{2-13}

        &BSR&CVPR'24&37.68&34.61&39.55&37.20&25.50&35.28&25.79&42.20&28.87*&34.08 \\
        &\cellcolor{gray!20}BSR-SA&\cellcolor{gray!20}Ours&\cellcolor{gray!20}\textbf{14.47}&\cellcolor{gray!20}\textbf{18.20}&\cellcolor{gray!20}\textbf{19.23}&\cellcolor{gray!20}\textbf{16.37}&\cellcolor{gray!20}\textbf{7.06}&\cellcolor{gray!20}\textbf{14.11}&\cellcolor{gray!20}\textbf{7.73}&\cellcolor{gray!20}\textbf{28.83}&\cellcolor{gray!20}\textbf{3.35}*&\cellcolor{gray!20}\textbf{14.37} \\
        \cmidrule{2-13}

        &I-C&MM'25&39.73&37.32&41.14&37.46&25.70&28.50&29.67&42.35&34.39*&36.25 \\
        &\cellcolor{gray!20}I-C-SA&\cellcolor{gray!20}Ours&\cellcolor{gray!20}\textbf{20.57}&\cellcolor{gray!20}\textbf{22.64}&\cellcolor{gray!20}\textbf{23.13}&\cellcolor{gray!20}\textbf{20.54}&\cellcolor{gray!20}\textbf{10.61}&\cellcolor{gray!20}\textbf{20.50}&\cellcolor{gray!20}\textbf{14.11}&\cellcolor{gray!20}\textbf{31.74}&\cellcolor{gray!20}\textbf{10.05}*&\cellcolor{gray!20}\textbf{19.32} \\
        \bottomrule
    \end{tabular}
    
    \label{table: experiment_untargeted_cityscapes}
\end{table*}

\begin{figure*}
    \centering
    \includegraphics[width=0.95\linewidth]{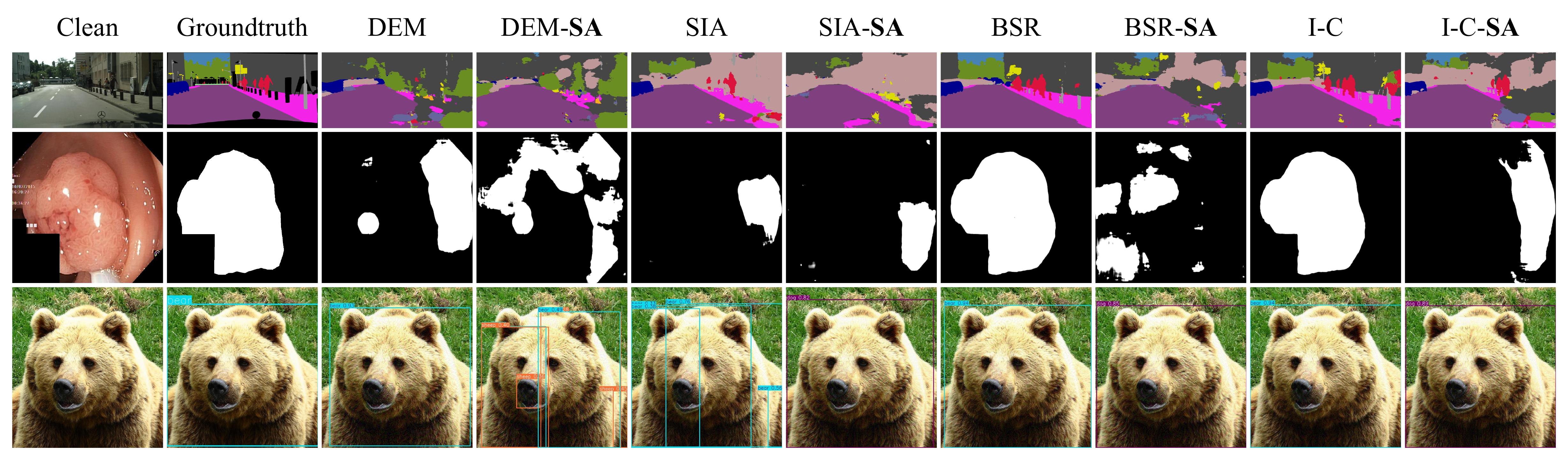}
    \caption{Visualized non-targeted results of diverse transformation-based attacks with and without spatial alignment.}
    \label{fig:visualization_nontargeted_attack}
\end{figure*}

\begin{figure*}
        \centering
    \includegraphics[width=0.95\linewidth]{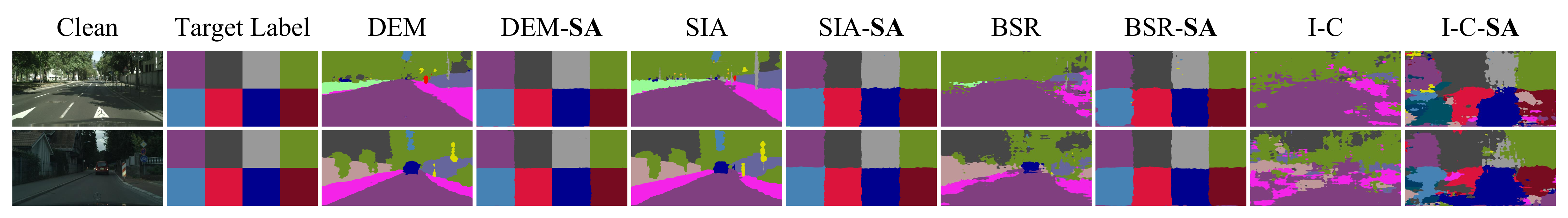}
    \caption{Visualized targeted results of diverse transformation-based attacks with and without spatial alignment.}
    \label{fig:visualization_targeted_attack}
\end{figure*}

\subsection{Spatial Alignment}
\label{section:Spatial Alignment}
Therefore, to ensure that $T(\bm{x}_{adv}^{(t-1)})$ and the label are spatially aligned, it is natural to apply the spatial transformation $T_{s}$ to the label $\bm{y}$ synchronously. Accordingly, to extend TAAs to structured tasks, we adjust Equation~\ref{equation:definition_TAA_1} as
\begin{equation}
\begin{aligned}
    \bar{\bm{\mu}}=\frac{1}{N}\Sigma_{i=1}^{N}\nabla_{\bm{x}_{adv}^{(t-1)}}J(f_{s}(T(\bm{x}_{adv}^{(t-1)})),\mathcolor{mdarkred}{T_{s}(\bm{y})}).
\end{aligned}
\label{equation:definition_spatial_alignment}
\end{equation}
The difference between Equation~\ref{equation:definition_spatial_alignment} and Equation~\ref{equation:definition_TAA_1} is that \textit{Spatial Alignment (SA)} is realized by applying $T_{s}$ to the label $\bm{y}$. We integrate SA into the general TAA framework and establish a unified \textit{Spatial Alignment Framework (SAF)} (detailed in Figure~\ref{fig:illustration_misalignment_alignment} and Algorithm~\ref{algorithm: Spatial Alignment}) for structured tasks.

The implementations of SA differ depending on the structured task, as shown in Figure~\ref{fig:illustration_misalignment_alignment}(b). For semantic segmentation, the label map is directly transformed alongside $\bm{x}_{adv}^{(t-1)}$ (e.g., a $\tau^{\circ}$ rotation of $\bm{x}_{adv}^{(t-1)}$ is applied in sync to the label map). For object detection, each bounding box's corners are recomputed according to the transformation. For example, if $\bm{x}_{adv}^{(t-1)}$ is divided into shuffled blocks by $T_{s}$, the coordinates of the bounding boxes are recalculated according to shuffling. If a block boundary intersects a box, the box may be split into multiple blocks to maintain proper alignment.

\section{Experiments and Results}

\subsection{Setup}
\textbf{Datasets.} For semantic segmentation, we employ $500$ Cityscapes~\cite{cordts2016cityscapes} (natural images with road scenes) validation images and $200$ Kvasir-SEG~\cite{jha2019kvasir} images (medical images for polyp segmentation) as the victim set, enabling validation across diverse domains. For object detection, we employ $5000$ MS COCO~\cite{lin2014microsoft} validation images as the victim set. Images from Cityscapes, Kvasir-SEG, and MS COCO are resized to $768\times768$, $448\times448$, and $640\times640$, respectively, to evaluate spatial alignment across different resolutions.

\textbf{Models.} We employ diverse models for comprehensive evaluations. Specifically, on Cityscapes, we consider DeepLabV3~\cite{chen2017rethinking}, DeepLabV3+~\cite{chen2018encoder}, FPN~\cite{lin2017feature}, PSPNet~\cite{zhao2017pyramid}, U-Net++~\cite{zhou2018unet++}, UPerNet~\cite{xiao2018unified}, DPT~\cite{ranftl2021vision}, and SegFormer~\cite{xie2021segformer}. On Kvasir-SEG, we consider ATT-UNet~\cite{oktay2018attention}, HRNet-18~\cite{wang2020deep}, I$^{2}$U-Net~\cite{dai2024i2u}, ResU-Net++~\cite{jha2019resunet++}, ResU-Net~\cite{diakogiannis2020resunet}, CCT-UNet~\cite{yan2023cct}, U-Net~\cite{ronneberger2015u}, U-Net++~\cite{zhou2018unet++}, and U-Net-URPC~\cite{xia2020uncertainty}. On MS COCO, we employ SSD~\cite{liu2016ssd}, Faster R-CNN~\cite{ren2015faster}, FCOS~\cite{tian2019fcos}, RetinaNet~\cite{lin2017focal}, YOLOv3~\cite{redmon2018yolov3}, YOLOv5~\cite{yolov5}, YOLOv8~\cite{yolov8_ultralytics}, YOLOv10~\cite{wang2024yolov10}, and YOLOv12~\cite{tian2025yolov12}.

All Cityscapes models are trained for $200$ epochs using the Semantic Segmentation Library~\cite{Iakubovskii:2019} with SGD (base learning rate $1\times10^{-2}$, batch size $16$, momentum $0.9$, weight decay $1\times10^{-4}$) and a polynomial learning rate schedule (power $0.9$). Data augmentation includes noise addition, rotation, resizing, and horizontal flipping. All Kvasir-SEG models are trained for $200$ epochs using the XNet~\cite{Zhou_2023_ICCV} repository with Adam (base learning rate $1\times10^{-3}$) and a polynomial learning rate schedule (power $0.9$). The batch size is $16$, except for HRNet-18, which uses a batch size of $8$. Data augmentation is identical to that used for Cityscapes. In object detection experiments, the YOLO-series models and RT-DETR are sourced from Ultralytics, whereas all other detectors are from torchvision.

\textbf{Baselines.} We evaluate spatial alignment across various TAAs, including DEM~\cite{zou2020improving} (ECCV'20), SIA~\cite{wang2023structure} (ICCV'23), BSR~\cite{wang2024boosting} (CVPR'24), and I-C attack~\cite{liang2025ic} (ACM MM'25). These attacks are implemented in accordance with their official specifications.

\textbf{Metrics.} For semantic segmentation, we adopt mIoU as the evaluation metric, while for object detection, we use mAP@50:95. In non-targeted attacks, lower values of mIoU and mAP suggest stronger attack effectiveness, while the opposite holds for targeted attacks.

\subsection{Non-Targeted Attacks on Semantic Segmentation}
\label{section: Non-Targeted Attacks on Semantic Segmentation}
This experiment compares the non-targeted attack performance of various TAAs with traditional framework and our SAF in semantic segmentation tasks. The perturbation budget $\epsilon=10/255$, the number of iterations $L=10$, and the step size $\alpha=1/255$. For Cityscapes, the surrogate models include DeepLabV3 (R50) and SegFormer (MiT-B2). For Kvasir-SEG, the surrogate models include U-Net and U-Net-URPC. In addition, SegPGD~\cite{gu2022segpgd} and CosPGD~\cite{agnihotri2024cospgd}, the state-of-the-art segmentation attacks, are used for comparison in Cityscapes.

The results are reported in Table~\ref{table: experiment_untargeted_cityscapes} and Table~\ref{table: experiment_untargeted_kvasir_seg}. The official SIA, BSR, and I-C employ large-magnitude spatial transformations that induce severe spatial misalignment, resulting in poor performance on semantic segmentation tasks. Our SAF effectively eliminates this misalignment, enabling SIA, BSR, and I-C to attack semantic segmentation models effectively. For DEM, the spatial transformations it adopts are relatively mild, which induces limited spatial misalignment. Consequently, DEM already exhibit reasonable performance even with traditional framework. However, our SAF remains beneficial, as it mitigates the residual misalignment and consistently yields significant gains. 

Moreover, an interesting finding is that the performance ranking of TAAs in semantic segmentation markedly differs from that in classification. While I-C attack is usually the strongest in classification, results on Cityscapes and Kvasir-SEG show a different or even opposite trend. We believe it reflects underlying insights that warrant future exploration.

\subsection{Non-Targeted Attacks on Object Detection}
\label{section: Non-Targeted Attacks on Object Detection}
This experiment compares the performance of non-targeted attacks in object detection with the same hyperparameters as in the semantic segmentation experiments. The surrogate models include YOLOv5-s and YOLOv8-n. The results reported in Table~\ref{table: experiment_untargeted_coco} highlight the importance of our SAF for effectively applying TAAs in object detection. 
Spatial alignment eliminates this misalignment and substantially boosts their performance.

\begin{table*}[t]
    \centering
    \fontsize{9}{10}\selectfont
    \setlength{\tabcolsep}{1.8mm}
    \caption{mIoU (\%) of non-targeted attacks on Kvasir-SEG. * indicates white-box attacks. -SA denotes attacks with spatial alignment.}
    \begin{tabular}{cllcccccccccc}
         \toprule
         Surrogate&Attack&Venue&ATTU&HRN18&I$^{2}$UN&ResU++&ResU&CCT-U&UNet&UN++&URPC&AVG \\
         \midrule
         
         \multicolumn{3}{c}{Clean}&85.05&84.11&86.81&74.34&74.54&83.55&83.35&83.79&83.12&82.07 \\
         \cmidrule{1-13}

        \multirow{8}{*}[-2ex]{UNet}&DEM&ECCV'20&16.86&18.96&66.91&63.14&23.14&18.24&13.66*&16.45&17.33&28.30\\
        &\cellcolor{gray!20}DEM-SA&\cellcolor{gray!20}Ours&\cellcolor{gray!20}\textbf{15.88}&\cellcolor{gray!20}\textbf{18.10}&\cellcolor{gray!20}\textbf{62.15}&\cellcolor{gray!20}\textbf{62.50}&\cellcolor{gray!20}\textbf{22.46}&\cellcolor{gray!20}\textbf{17.25}&\cellcolor{gray!20}\textbf{12.51}*&\cellcolor{gray!20}\textbf{15.58}&\cellcolor{gray!20}\textbf{16.23}&\cellcolor{gray!20}\textbf{26.96}\\
        \cmidrule{2-13}
         &SIA&ICCV'23&31.53&32.59&70.95&65.62&32.52&32.72&26.53*&31.36&32.33&39.57\\

         &\cellcolor{gray!20}SIA-SA&\cellcolor{gray!20}Ours&\cellcolor{gray!20}\textbf{19.65}&\cellcolor{gray!20}\textbf{20.07}&\cellcolor{gray!20}\textbf{50.92}&\cellcolor{gray!20}\textbf{59.10}&\cellcolor{gray!20}\textbf{21.87}&\cellcolor{gray!20}\textbf{20.17}&\cellcolor{gray!20}\textbf{17.74}*&\cellcolor{gray!20}\textbf{19.60}&\cellcolor{gray!20}\textbf{20.31}&\cellcolor{gray!20}\textbf{27.71}\\
         \cmidrule{2-13}

        &BSR&CVPR'24&61.15&60.76&78.55&72.48&55.83&61.48&57.72*&60.58&61.43&63.33\\
        &\cellcolor{gray!20}BSR-SA&\cellcolor{gray!20}Ours&\cellcolor{gray!20}\textbf{20.92}&\cellcolor{gray!20}\textbf{22.13}&\cellcolor{gray!20}\textbf{57.37}&\cellcolor{gray!20}\textbf{62.81}&\cellcolor{gray!20}\textbf{25.34}&\cellcolor{gray!20}\textbf{21.73}&\cellcolor{gray!20}\textbf{18.23}*&\cellcolor{gray!20}\textbf{20.86}&\cellcolor{gray!20}\textbf{21.89}&\cellcolor{gray!20}\textbf{30.14}\\
        \cmidrule{2-13}

        &I-C&MM'25&54.59&53.90&79.99&70.75&49.81&55.01&49.60*&53.67&54.69&58.00\\
        &\cellcolor{gray!20}I-C-SA&\cellcolor{gray!20}Ours&\cellcolor{gray!20}\textbf{26.99}&\cellcolor{gray!20}\textbf{27.30}&\cellcolor{gray!20}\textbf{65.57}&\cellcolor{gray!20}\textbf{63.87}&\cellcolor{gray!20}\textbf{27.56}&\cellcolor{gray!20}\textbf{27.78}&\cellcolor{gray!20}\textbf{22.56}*&\cellcolor{gray!20}\textbf{26.65}&\cellcolor{gray!20}\textbf{27.99}&\cellcolor{gray!20}\textbf{35.14}\\
        \midrule

        \multirow{8}{*}[-2ex]{U-URPC}&DEM&ECCV'20&16.81&19.33&72.65&64.56&23.20&17.33&17.57&19.06&15.62*&29.57\\
        &\cellcolor{gray!20}DEM-SA&\cellcolor{gray!20}Ours&\cellcolor{gray!20}\textbf{14.90}&\cellcolor{gray!20}\textbf{17.76}&\cellcolor{gray!20}\textbf{67.53}&\cellcolor{gray!20}\textbf{63.44}&\cellcolor{gray!20}\textbf{21.62}&\cellcolor{gray!20}\textbf{15.39}&\cellcolor{gray!20}\textbf{15.55}&\cellcolor{gray!20}\textbf{17.26}&\cellcolor{gray!20}\textbf{13.30}*&\cellcolor{gray!20}\textbf{27.42}\\
        \cmidrule{2-13}
        
         &SIA&ICCV'23&41.65&41.88&75.49&67.46&41.96&42.17&42.50&43.41&36.81*&48.15\\
         &\cellcolor{gray!20}SIA-SA&\cellcolor{gray!20}Ours&\cellcolor{gray!20}\textbf{23.44}&\cellcolor{gray!20}\textbf{23.50}&\cellcolor{gray!20}\textbf{56.54}&\cellcolor{gray!20}\textbf{60.14}&\cellcolor{gray!20}\textbf{25.73}&\cellcolor{gray!20}\textbf{24.28}&\cellcolor{gray!20}\textbf{24.73}&\cellcolor{gray!20}\textbf{25.12}&\cellcolor{gray!20}\textbf{22.60}*&\cellcolor{gray!20}\textbf{31.79}\\
         \cmidrule{2-13}

        &BSR&CVPR'24&63.89&64.01&80.51&73.21&56.82&64.43&64.24&64.82&63.03*&66.11\\
        &\cellcolor{gray!20}BSR-SA&\cellcolor{gray!20}Ours&\cellcolor{gray!20}\textbf{20.89}&\cellcolor{gray!20}\textbf{21.53}&\cellcolor{gray!20}\textbf{60.00}&\cellcolor{gray!20}\textbf{63.54}&\cellcolor{gray!20}\textbf{25.69}&\cellcolor{gray!20}\textbf{21.24}&\cellcolor{gray!20}\textbf{21.97}&\cellcolor{gray!20}\textbf{22.41}&\cellcolor{gray!20}\textbf{19.78}*&\cellcolor{gray!20}\textbf{30.78}\\
        \cmidrule{2-13}

        &I-C&MM'25&63.65&63.27&82.76&72.52&58.30&64.25&63.43&64.68&63.04*&66.21\\
        &\cellcolor{gray!20}I-C-SA&\cellcolor{gray!20}Ours&\cellcolor{gray!20}\textbf{37.36}&\cellcolor{gray!20}\textbf{37.97}&\cellcolor{gray!20}\textbf{69.62}&\cellcolor{gray!20}\textbf{64.80}&\cellcolor{gray!20}\textbf{38.35}&\cellcolor{gray!20}\textbf{38.26}&\cellcolor{gray!20}\textbf{38.39}&\cellcolor{gray!20}\textbf{39.32}&\cellcolor{gray!20}\textbf{36.12}*&\cellcolor{gray!20}\textbf{44.47}\\
        \bottomrule
    \end{tabular}
    
    \label{table: experiment_untargeted_kvasir_seg}
\end{table*}

\begin{table}[t]
    \centering
    \caption{mIoU (\%) of targeted attacks on Cityscapes.}
    \fontsize{9}{10}\selectfont
    \setlength{\tabcolsep}{1.6mm}
    \begin{tabular}{ccrrrrr}
    \toprule
         &&DEM&SIA&BSR&I-C&\textbf{AVG} \\
         \midrule
         \multirow{2}{*}[0ex]{\tabincell{c}{DLv3\\R50}}&Official&1.71&1.68&1.77&1.86&1.76\\
         &\cellcolor{gray!20}Official-SA&\cellcolor{gray!20}\textbf{52.39}&\cellcolor{gray!20}\textbf{43.03}&\cellcolor{gray!20}\textbf{42.53}&\cellcolor{gray!20}\textbf{22.79}&\cellcolor{gray!20}\textbf{40.19}\\
         \midrule
         \multirow{2}{*}[0ex]{\tabincell{c}{FPN\\R50}}&Official&1.56&1.65&1.66&1.70&1.64\\
         &\cellcolor{gray!20}Official-SA&\cellcolor{gray!20}\textbf{14.46}&\cellcolor{gray!20}\textbf{20.65}&\cellcolor{gray!20}\textbf{15.75}&\cellcolor{gray!20}\textbf{7.51}&\cellcolor{gray!20}\textbf{14.59}\\
         
    \bottomrule
    \end{tabular}
    \label{table:targeted_attacks_on_cityscapes}
\end{table}

\subsection{Targeted Attacks on Cityscapes}
The preceding experiments demonstrate the importance of SAF for TAAs under non-targeted attacks. This experiment further shows that SAF plays an even more critical role for TAAs in targeted attacks. We conduct white-box targeted attacks on DeepLabV3 (R50) on Cityscapes. The perturbation budget $\epsilon=16/255$, the number of iterations $L=200$, and the step size $\alpha=0.08/255$. The target label is a map composed of eight semantic block. Quantitative results are reported in Table~\ref{table:targeted_attacks_on_cityscapes}, and visualizations are shown in Figure~\ref{fig:visualization_targeted_attack}. It can be observed that targeted attacks are more sensitive to spatial misalignment. Consequently, TAAs with traditional framework almost completely fail under the targeted scenarios. In contrast, SAF restores their effectiveness.

\begin{table}[t]
    \centering
    \caption{Results of non-targeted TAAs with and without SA under input preprocessing defenses. \textit{w/o} refers to results without defense.}
    \fontsize{8}{10}\selectfont
    \setlength{\tabcolsep}{0.7mm}
    \begin{tabular}{lrrrrrrrrr}
        \toprule
            &\multicolumn{3}{c}{Cityscapes}&\multicolumn{3}{c}{Kvasir-SEG}&\multicolumn{3}{c}{MS COCO}\\
            \cmidrule{2-4}\cmidrule{5-7}\cmidrule{8-10}
            &w/o&JPEG&BitR&w/o&JPEG&BitR&w/o&JPEG&BitR\\
            \midrule
            DEM&8.25&8.79&8.22&28.30&28.64&28.18&10.82&11.62&10.86\\
            \cellcolor{gray!20}DEM-SA&\cellcolor{gray!20}\textbf{6.59}&\cellcolor{gray!20}\textbf{6.98}&\cellcolor{gray!20}\textbf{6.57}&\cellcolor{gray!20}\textbf{26.96}&\cellcolor{gray!20}\textbf{27.38}&\cellcolor{gray!20}\textbf{26.83}&\cellcolor{gray!20}\textbf{3.85}&\cellcolor{gray!20}\textbf{4.32}&\cellcolor{gray!20}\textbf{3.87}\\
            \midrule
            
            SIA&14.20&15.77&14.23&39.57&39.72&39.56&12.18&13.05&12.22\\
            \cellcolor{gray!20}SIA-SA&\cellcolor{gray!20}\textbf{6.61}&\cellcolor{gray!20}\textbf{7.60}&\cellcolor{gray!20}\textbf{6.63}&\cellcolor{gray!20}\textbf{27.71}&\cellcolor{gray!20}\textbf{27.90}&\cellcolor{gray!20}\textbf{27.70}&\cellcolor{gray!20}\textbf{2.85}&\cellcolor{gray!20}\textbf{3.42}&\cellcolor{gray!20}\textbf{2.87}\\
            \midrule
            
            BSR&39.96&40.35&39.97&63.33&63.43&63.32&21.08&22.05&21.13\\
            \cellcolor{gray!20}BSR-SA&\cellcolor{gray!20}\textbf{10.14}&\cellcolor{gray!20}\textbf{11.19}&\cellcolor{gray!20}\textbf{10.17}&\cellcolor{gray!20}\textbf{30.14}&\cellcolor{gray!20}\textbf{30.30}&\cellcolor{gray!20}\textbf{30.13}&\cellcolor{gray!20}\textbf{5.60}&\cellcolor{gray!20}\textbf{6.31}&\cellcolor{gray!20}\textbf{5.63}\\
            \midrule
            
            I-C&30.29&30.95&30.25&58.00&58.12&57.98&27.16&27.78&22.64\\
            \cellcolor{gray!20}I-C-SA&\cellcolor{gray!20}\textbf{14.06}&\cellcolor{gray!20}\textbf{15.07}&\cellcolor{gray!20}\textbf{14.05}&\cellcolor{gray!20}\textbf{35.14}&\cellcolor{gray!20}\textbf{35.34}&\cellcolor{gray!20}\textbf{35.12}&\cellcolor{gray!20}\textbf{4.81}&\cellcolor{gray!20}\textbf{5.44}&\cellcolor{gray!20}\textbf{4.83}\\
        \bottomrule
    \end{tabular}

    \label{table:Input Preprocessing Defense}
\end{table}

\subsection{Attacks under Input Preprocessing Defense}
In this experiment, we further examine whether SAF-based TAAs remain more effective than their non-SAF counterparts under defensive scenarios. Two classical defenses, JPEG compression~\cite{das2017keeping} and bit-depth reduction~\cite{guo2017countering}, are considered. The JPEG quality factor is set to $90$, and the bit-depth reduction uses $6$ bits. We evaluate the non-targeted adversarial examples generated in Section~\ref{section: Non-Targeted Attacks on Semantic Segmentation} and~\ref{section: Non-Targeted Attacks on Object Detection}, with DeepLabV3 (R50), U-Net, and YOLOv5 as surrogate models for Cityscapes, Kvasir-SEG and MS COCO, respectively. As shown in Table~\ref{table:Input Preprocessing Defense}, these basic defenses have limited impact on TAA-generated adversarial examples, while SAF-based TAAs consistently achieve stronger attack performance.

\begin{table*}[t]
    \centering
    \fontsize{9}{10}\selectfont
    \setlength{\tabcolsep}{1.2mm}
    \caption{mAP@50:95 (\%) of non-targeted attacks on MS COCO. * indicates white-box attacks. -SA denotes attacks with spatial alignment.}
    \begin{tabular}{cllccccccccccc}
         \toprule
         \multirow{2}{*}[0ex]{Surrogate}&\multirow{2}{*}[0ex]{Attack}&\multirow{2}{*}[0ex]{Venue}&\multicolumn{1}{c}{YLV3}&\multicolumn{1}{c}{YLV5}&\multicolumn{1}{c}{YLV8}&\multicolumn{1}{c}{YLV10}&\multicolumn{1}{c}{YLV12}&\multicolumn{1}{c}{SSD}&\multicolumn{1}{c}{FRCNN}&\multicolumn{1}{c}{FCOS}&\multicolumn{1}{c}{Retina}&\multicolumn{1}{c}{RT-DETR}&\multirow{2}{*}[0ex]{AVG} \\
         &&&\multicolumn{1}{c}{-u}&\multicolumn{1}{c}{-s}&\multicolumn{1}{c}{-n}&\multicolumn{1}{c}{-x}&\multicolumn{1}{c}{-l}&\multicolumn{1}{c}{VGG16}&\multicolumn{1}{c}{R50}&\multicolumn{1}{c}{R50}&\multicolumn{1}{c}{R50}&\multicolumn{1}{c}{-l}& \\
        \midrule
        \multicolumn{3}{c}{Clean}&48.07&38.85&33.83&51.32&50.85&24.92&33.70&35.33&32.73&51.90&40.15 \\
         \cmidrule{1-14}

        \multirow{8}{*}[-2ex]{\tabincell{c}{YLV5\\-s}}&DEM&ECCV'20&10.55&2.04*&5.15&13.61&15.93&16.16&8.85&8.98&9.65&17.32&10.82 \\
        &\cellcolor{gray!20}DEM-SA&\cellcolor{gray!20}Ours&\cellcolor{gray!20}\textbf{2.85}&\cellcolor{gray!20}\textbf{0.08}*&\cellcolor{gray!20}\textbf{0.42}&\cellcolor{gray!20}\textbf{5.04}&\cellcolor{gray!20}\textbf{6.90}&\cellcolor{gray!20}\textbf{9.77}&\cellcolor{gray!20}\textbf{1.72}&\cellcolor{gray!20}\textbf{1.94}&\cellcolor{gray!20}\textbf{2.21}&\cellcolor{gray!20}\textbf{7.60}&\cellcolor{gray!20}\textbf{3.85} \\
        \cmidrule{2-14}
        
         &SIA&ICCV'23&12.67&1.57*&6.86&15.03&16.11&17.23&10.73&10.86&11.44&19.28&12.18 \\
         &\cellcolor{gray!20}SIA-SA&\cellcolor{gray!20}Ours&\cellcolor{gray!20}\textbf{1.58}&\cellcolor{gray!20}\textbf{0.03}*&\cellcolor{gray!20}\textbf{0.37}&\cellcolor{gray!20}\textbf{2.59}&\cellcolor{gray!20}\textbf{3.10}&\cellcolor{gray!20}\textbf{10.59}&\cellcolor{gray!20}\textbf{1.31}&\cellcolor{gray!20}\textbf{1.49}&\cellcolor{gray!20}\textbf{1.66}&\cellcolor{gray!20}\textbf{5.73}&\cellcolor{gray!20}\textbf{2.85} \\
         \cmidrule{2-14}

        &BSR&CVPR'24&23.14&11.08*&16.58&25.73&26.60&20.69&19.14&19.13&19.54&29.15&21.08 \\
        &\cellcolor{gray!20}BSR-SA&\cellcolor{gray!20}Ours&\cellcolor{gray!20}\textbf{4.62}&\cellcolor{gray!20}\textbf{0.71}*&\cellcolor{gray!20}\textbf{2.36}&\cellcolor{gray!20}\textbf{6.36}&\cellcolor{gray!20}\textbf{7.33}&\cellcolor{gray!20}\textbf{12.05}&\cellcolor{gray!20}\textbf{3.79}&\cellcolor{gray!20}\textbf{3.91}&\cellcolor{gray!20}\textbf{4.43}&\cellcolor{gray!20}\textbf{10.40}&\cellcolor{gray!20}\textbf{5.60} \\
        \cmidrule{2-14}

        &I-C&MM'25&29.54&19.76*&23.01&32.52&33.39&22.22&24.75&24.89&24.73&36.82&27.16 \\
        &\cellcolor{gray!20}I-C-SA&\cellcolor{gray!20}Ours&\cellcolor{gray!20}\textbf{3.44}&\cellcolor{gray!20}\textbf{0.79}*&\cellcolor{gray!20}\textbf{1.99}&\cellcolor{gray!20}\textbf{4.95}&\cellcolor{gray!20}\textbf{5.48}&\cellcolor{gray!20}\textbf{11.04}&\cellcolor{gray!20}\textbf{3.40}&\cellcolor{gray!20}\textbf{3.53}&\cellcolor{gray!20}\textbf{3.75}&\cellcolor{gray!20}\textbf{9.75}&\cellcolor{gray!20}\textbf{4.81} \\
        
         \midrule

        \multirow{8}{*}[-2ex]{\tabincell{c}{YLV8\\-n}}&DEM&ECCV'20&14.76&4.53&1.42*&19.50&21.82&14.18&9.33&9.57&10.30&20.88&12.63 \\
        &\cellcolor{gray!20}DEM-SA&\cellcolor{gray!20}Ours&\cellcolor{gray!20}\textbf{8.26}&\cellcolor{gray!20}\textbf{0.60}&\cellcolor{gray!20}\textbf{0.05}*&\cellcolor{gray!20}\textbf{12.53}&\cellcolor{gray!20}\textbf{14.85}&\cellcolor{gray!20}\textbf{8.25}&\cellcolor{gray!20}\textbf{3.23}&\cellcolor{gray!20}\textbf{3.63}&\cellcolor{gray!20}\textbf{3.88}&\cellcolor{gray!20}\textbf{13.58}&\cellcolor{gray!20}\textbf{6.89} \\
        \cmidrule{2-14}
         &SIA&ICCV'23&15.19&5.12&0.78*&18.58&18.87&15.70&10.39&10.54&11.15&21.65&12.80 \\
         &\cellcolor{gray!20}SIA-SA&\cellcolor{gray!20}Ours&\cellcolor{gray!20}\textbf{4.94}&\cellcolor{gray!20}\textbf{0.41}&\cellcolor{gray!20}\textbf{0.01}*&\cellcolor{gray!20}\textbf{7.94}&\cellcolor{gray!20}\textbf{8.36}&\cellcolor{gray!20}\textbf{9.75}&\cellcolor{gray!20}\textbf{2.33}&\cellcolor{gray!20}\textbf{2.66}&\cellcolor{gray!20}\textbf{2.79}&\cellcolor{gray!20}\textbf{11.04}&\cellcolor{gray!20}\textbf{5.02} \\
         \cmidrule{2-14}

        &BSR&CVPR'24&23.49&14.28&8.65*&27.12&27.64&18.81&17.83&17.86&18.25&28.95&20.29 \\
        &\cellcolor{gray!20}BSR-SA&\cellcolor{gray!20}Ours&\cellcolor{gray!20}\textbf{7.35}&\cellcolor{gray!20}\textbf{1.72}&\cellcolor{gray!20}\textbf{0.52}*&\cellcolor{gray!20}\textbf{10.67}&\cellcolor{gray!20}\textbf{11.90}&\cellcolor{gray!20}\textbf{10.04}&\cellcolor{gray!20}\textbf{3.94}&\cellcolor{gray!20}\textbf{4.38}&\cellcolor{gray!20}\textbf{4.69}&\cellcolor{gray!20}\textbf{13.47}&\cellcolor{gray!20}\textbf{6.87} \\
        \cmidrule{2-14}

        &I-C&MM'25&29.11&20.09&15.96*&33.94&34.60&21.43&22.99&23.53&23.17&36.87&26.17 \\
        &\cellcolor{gray!20}I-C-SA&\cellcolor{gray!20}Ours&\cellcolor{gray!20}\textbf{5.19}&\cellcolor{gray!20}\textbf{1.34}&\cellcolor{gray!20}\textbf{0.42}*&\cellcolor{gray!20}\textbf{8.89}&\cellcolor{gray!20}\textbf{9.48}&\cellcolor{gray!20}\textbf{10.95}&\cellcolor{gray!20}\textbf{3.78}&\cellcolor{gray!20}\textbf{3.14}&\cellcolor{gray!20}\textbf{4.35}&\cellcolor{gray!20}\textbf{13.15}&\cellcolor{gray!20}\textbf{6.07} \\

        \bottomrule
    \end{tabular}
    
    \label{table: experiment_untargeted_coco}
\end{table*}

\begin{figure}[t]
    \centering
    \includegraphics[width=1.00\linewidth]{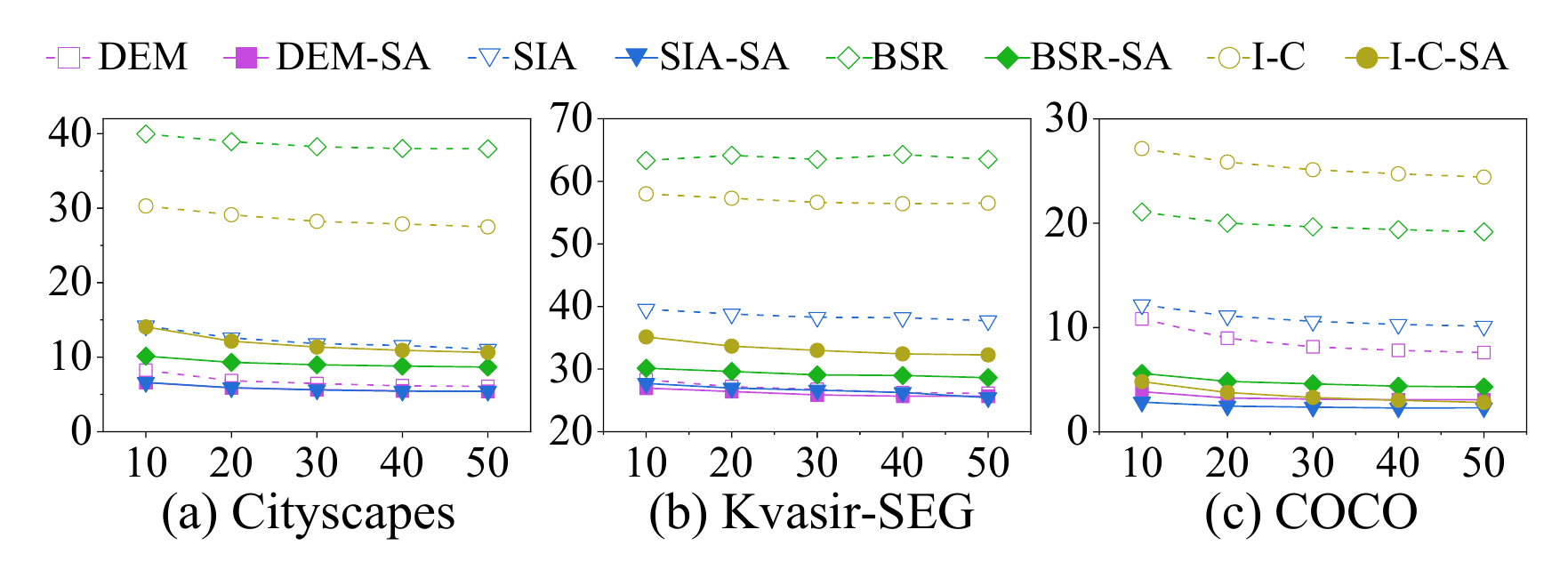}
    \caption{Ablation studies on the number of iterations $L$.}
    \label{fig:ablation_iterations}
\end{figure}

\subsection{Ablation Study on Iterations}
\label{section: Ablation Study on Iterations}
In this experiment, we conduct an ablation study on the number of iterations $L$ to investigate the effect of SAF on TAAs under different iteration settings. We perform the ablation on Cityscapes, Kvasir-SEG, and MS COCO with DeepLabV3 (R50), U-Net, and YOLOv5 as surrogate models, respectively. The perturbation budget is fixed to $\epsilon=10/255$, the number of iterations $L$ ranges from $10$ to $50$ with an interval of $10$, and the step size $\alpha=\epsilon/L$. Results presented in Figure~\ref{fig:ablation_iterations} show that attacks are close to convergence at $L=10$. Increasing the number of iterations leads to only marginal performance gains. Across different iteration settings, our SAF consistently provides a significant boost.

\begin{figure}[t]
    \centering
    \includegraphics[width=1.00\linewidth]{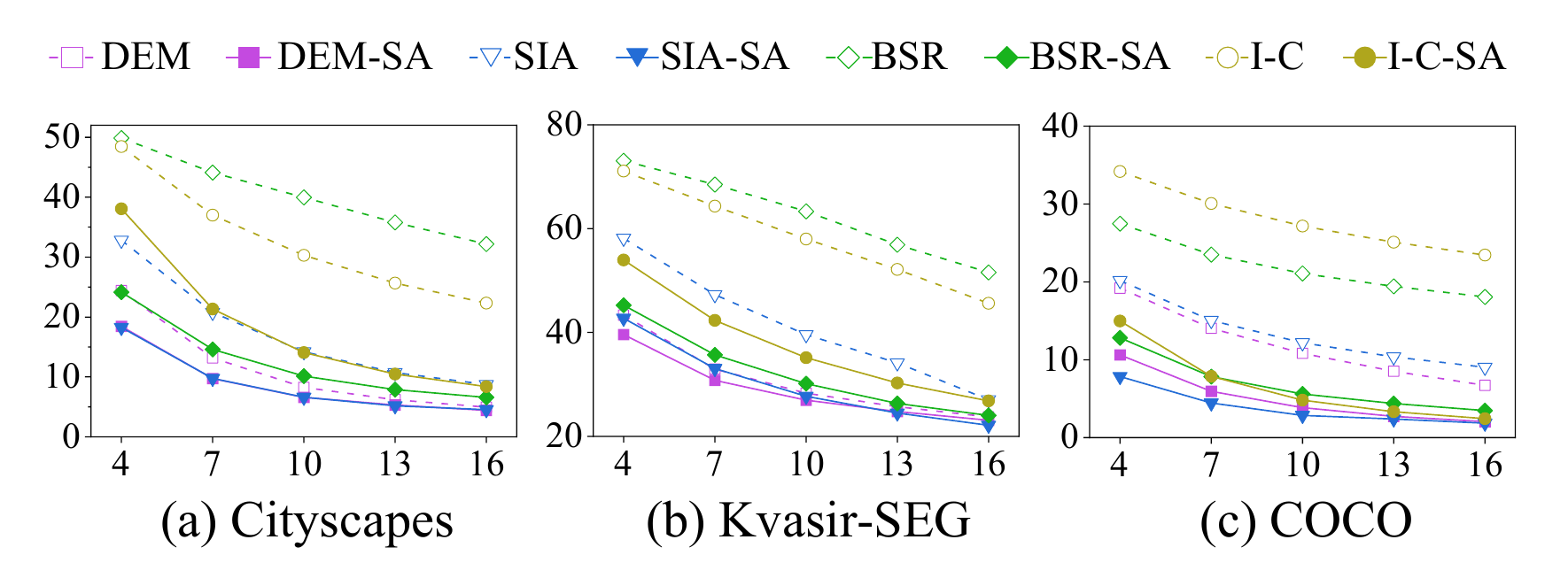}
    \caption{Ablation studies on perturbation budgets $\epsilon$.}
    \label{fig:ablation_budgets}
\end{figure}

\subsection{Ablation Study on Perturbation Budget}
Similarly, we conduct an ablation study on the perturbation budget $\epsilon$ to examine the effect of SAF on TAAs under different perturbation budgets. The experimental setup follows Section~\ref{section: Ablation Study on Iterations}, except that the number of iterations is fixed to $L=10$, and the perturbation budget $\epsilon$ ranges from $4/255$ to $16/255$ with an interval of $3/255$. The results are shown in Figure~\ref{fig:ablation_budgets}. As the perturbation budget increases, the attack performance improves steadily, and our SAF consistently enhances the performance of TAAs.

\section{Conclusions and Limitations}
Despite significant progress in non-structured tasks, TAAs perform worse or even fail when applied to structured tasks like semantic segmentation and object detection. In this work, we identify the root cause as spatial misalignment introduced by TAAs' spatial transformations, which leads to incorrect perturbations in structured tasks. Accordingly, we propose the \textit{Spatial Alignment} algorithm, which transforms structured labels synchronously with the input to effectively address this issue, enabling existing TAAs to be directly applied to structured tasks without modifying their original designs. 
Extensive experiments demonstrate that SA enables various TAAs to be effective on structured tasks. However, this work focuses on investigating the reasons behind the failure of TAAs on structured tasks and exploring how to make TAAs effective in these settings. We do not delve into further details, leaving many valuable directions for future research. For example, the mechanisms underlying the significant performance difference of TAAs between non-structured and structured tasks, as well as the differences in optimal transformation types and parameters for TAAs across these tasks, remain open questions. We look forward to future work in these promising directions.

\section*{Impact Statement}
This paper presents work whose goal is to advance the field of machine learning. There are many potential societal consequences of our work, none of which we feel must be specifically highlighted here.


\bibliography{example_paper}
\bibliographystyle{icml2026}




\end{document}